\documentclass[letterpaper, 10 pt, conference]{ieeeconf}
\pdfoutput=1  % for arXiv

\IEEEoverridecommandlockouts

\usepackage{booktabs}
\usepackage{pifont}
\newcommand{\cmark}{\ding{51}}

\usepackage{tabularx}
\usepackage{nicefrac}
\usepackage{cite}
\usepackage[caption=false,font=footnotesize]{subfig}

\usepackage[pdftex]{graphicx}

\usepackage{amsmath}
\usepackage{amssymb}

\usepackage{hologo}
\usepackage{listings}
\usepackage{hyperref}

\title{\LARGE \bf
Towards Scenario- and Capability-Driven Dataset Development and Evaluation: An Approach in the Context of\\Mapless Automated Driving
}

\author{Felix Gr\"un, Marcus Nolte, and Markus Maurer% <-this % stops a space
\thanks{The authors are with the Institute of \mbox{Control} \mbox{Engineering}, TU Braunschweig, 38106 Braunschweig, Germany
{\tt\small \{gruen, nolte, maurer\}@ifr.ing.tu-bs.de}}% <-this % stops a space
}

\begin{document}

% for arXiv publication with appropriate copyright notice
\twocolumn[
\begin{@twocolumnfalse}
\Huge {IEEE copyright notice} \\ \\
\large {\copyright\ 2024 IEEE. Personal use of this material is permitted. Permission from IEEE must be obtained for all other uses, in any current or future media, including reprinting/republishing this material for advertising or promotional purposes, creating new collective works, for resale or redistribution to servers or lists, or reuse of any copyrighted component of this work in other works.} \\ \\

{\Large Accepted for publication at the \emph{2024 35th IEEE Intelligent Vehicles Symposium (IV)}, Jeju Island, Korea, June 2 - 5, 2024} \\ \\

{\Large For the final published paper see DOI: \href{https://doi.org/10.1109/IV55156.2024.10588871}{10.1109/IV55156.2024.10588871}} \\ \\

Cite as:

\vspace{0.1cm}
\noindent\fbox{%
    \parbox{\textwidth}{%
        F.~Gr\"un, M.~Nolte, and M.~Maurer, ``Towards Scenario- and Capability-Driven Dataset Development and Evaluation: An Approach in the Context of Mapless Automated Driving,''
   in \emph{2024 35th IEEE Intelligent Vehicles Symposium (IV)}, Jeju Island, Korea, 2024, pp. 2176--2183, DOI: {10.1109/IV55156.2024.10588871}.
    }%
}
\vspace{2cm}

\end{@twocolumnfalse}
]

\noindent\begin{minipage}{\textwidth}

\hologo{BibTeX}:
\footnotesize
\begin{lstlisting}[frame=single]
@inproceedings{gruen_dataset_development_2024,
  author={Gr\"un, Felix and Nolte, Marcus and Maurer, Markus},
  booktitle={2024 35th IEEE Intelligent Vehicles Symposium (IV)},
  title={Towards Scenario- and Capability-Driven Dataset Development and Evaluation: An Approach in the Context of Mapless Automated Driving},
  address={Jeju Island, Korea},
  year={2024},
  pages={2176--2183},
  doi={10.1109/IV55156.2024.10588871},
  publisher={IEEE}
}
\end{lstlisting}
\end{minipage}

\maketitle
\thispagestyle{empty}
\pagestyle{empty}

%%%%%%%%%%%%%%%%%%%%%%%%%%%%%%%%%%%%%%%%%%%%%%%%%%%%%%%%%%%%%%%%%%%%%%%%%%%%%%%%
\begin{abstract}
    The foundational role of datasets in defining the capabilities of deep learning models has led to their rapid proliferation. At the same time, published research focusing on the process of dataset development for environment perception in automated driving has been scarce, thereby reducing the applicability of openly available datasets and impeding the development of effective environment perception systems. Sensor-based, mapless automated driving is one of the contexts where this limitation is evident. While leveraging real-time sensor data, instead of pre-defined HD maps promises enhanced adaptability and safety by effectively navigating unexpected environmental changes, it also increases the demands on the scope and complexity of the information provided by the perception system.

    To address these challenges, we propose a scenario- and capability-based approach to dataset development. Grounded in the principles of ISO~21448 (safety of the intended functionality, SOTIF), extended by ISO/TR~4804, our approach facilitates the structured derivation of dataset requirements. This not only aids in the development of meaningful new datasets but also enables the effective comparison of existing ones. Applying this methodology to a broad range of existing lane detection datasets, we identify significant limitations in current datasets, particularly in terms of real-world applicability, a lack of labeling of critical features, and an absence of comprehensive information for complex driving maneuvers.

\end{abstract}

%%%%%%%%%%%%%%%%%%%%%%%%%%%%%%%%%%%%%%%%%%%%%%%%%%%%%%%%%%%%%%%%%%%%%%%%%%%%%%%%
\section{Introduction}

Deep learning has revolutionized perception technologies in automated driving, considerably advancing the interpretation of complex vehicle environments. This transformative impact is also evident in lane detection where the ability to merge semantic scene understanding with detailed image analysis has enabled the generation of highly accurate predictions in diverse and challenging real-world conditions~\cite{vpgnet:lee:2017,robustlanedetection:zou:2019,robustlanedetection:lee:2022}. This advancement, however, is heavily reliant on the availability of extensive, relevant training data, which has led to the creation of numerous lane detection datasets. Despite this progress, the development and documentation of these datasets lack a structured, standardized approach, leading to a fragmented landscape of methodologies in data collection and labeling. This affects the applicability of developed datasets to real-world driving scenarios, and hinders the assessment and comparison of datasets, posing a significant challenge to researchers and industry professionals alike. Furthermore, many existing surveys on lane detection datasets are limited in the number of considered datasets and narrowly focus on basic metrics such as dataset size, which do not reflect the quality or relevance of the data~\cite{lanedatasetssurvey:shirke:2019,safetodrive:guo:2019,ttlane:liang:2020,lanemarkingsurvey:zhang:2021}.

In the context of mapless driving, the role of perception systems expands beyond lane assist functions to encompass comprehensive environment modeling for automated driving tasks. This extended role of perception systems amplifies the need to understand the specific information required for subsequent driving functions as well as the optimal way to encode this data for training artificial neural networks. Addressing these needs requires a coordinated effort from experts in machine learning, dataset design, and engineering of driving functions. The challenges include determining the essential information to be extracted from the vehicle's environment, developing labeling specifications to minimize ambiguities, and managing edge cases in input data.

Recognizing these challenges, Annex B of ISO/TR~4804~\cite{isotr4804:iso:2020} provides a framework for dataset construction, emphasizing the need for datasets to match attribute requirements defined in the initial stages of the dataset development process. It provides lists of example considerations, guiding questions, and typical artifacts of what it calls the `Define phase'~\cite[Annex~B]{isotr4804:iso:2020}. In addition to this, ISO/TR4804 formulates basic capabilities necessary for automated driving systems to fulfill the dynamic driving task. However, the report stops short of discussing methods for the derivation of these requirements. We are approaching these challenges from the perspective of ISO~21448 (SOTIF)~\cite{iso21448:iso:2022} compliant development processes for automated vehicles and propose the use of a scenario- and capability-based development process~\cite{skillbaseddevelopment:nolte:2017}. This approach aligns the development process with real-world scenarios and the specific capabilities required by the vehicle, providing a method to focus on the behaviors a vehicle must exhibit to complete driving tasks rather than the technical processes behind them. This abstract description facilitates the identification of gaps in the capabilities of the vehicle, reducing the likelihood of overlooking vital safety-related considerations and ensuring that the vehicle is equipped to handle a wide range of driving scenarios. In previous work this approach has been applied to derive requirements for automated driving systems (ADS)~\cite{skillbaseddevelopment:nolte:2017}. In this paper, we explore how it can help determine the requirements for training and testing datasets early in the development process, and how these can be used to aid in the meaningful comparison of existing datasets.

\setlength{\tabcolsep}{0pt}
\begin{figure*}[ht!]
    \centering
    \begin{tabular}{cc}
        \subfloat[Scenario 1)]{\includegraphics[width=0.48\textwidth]{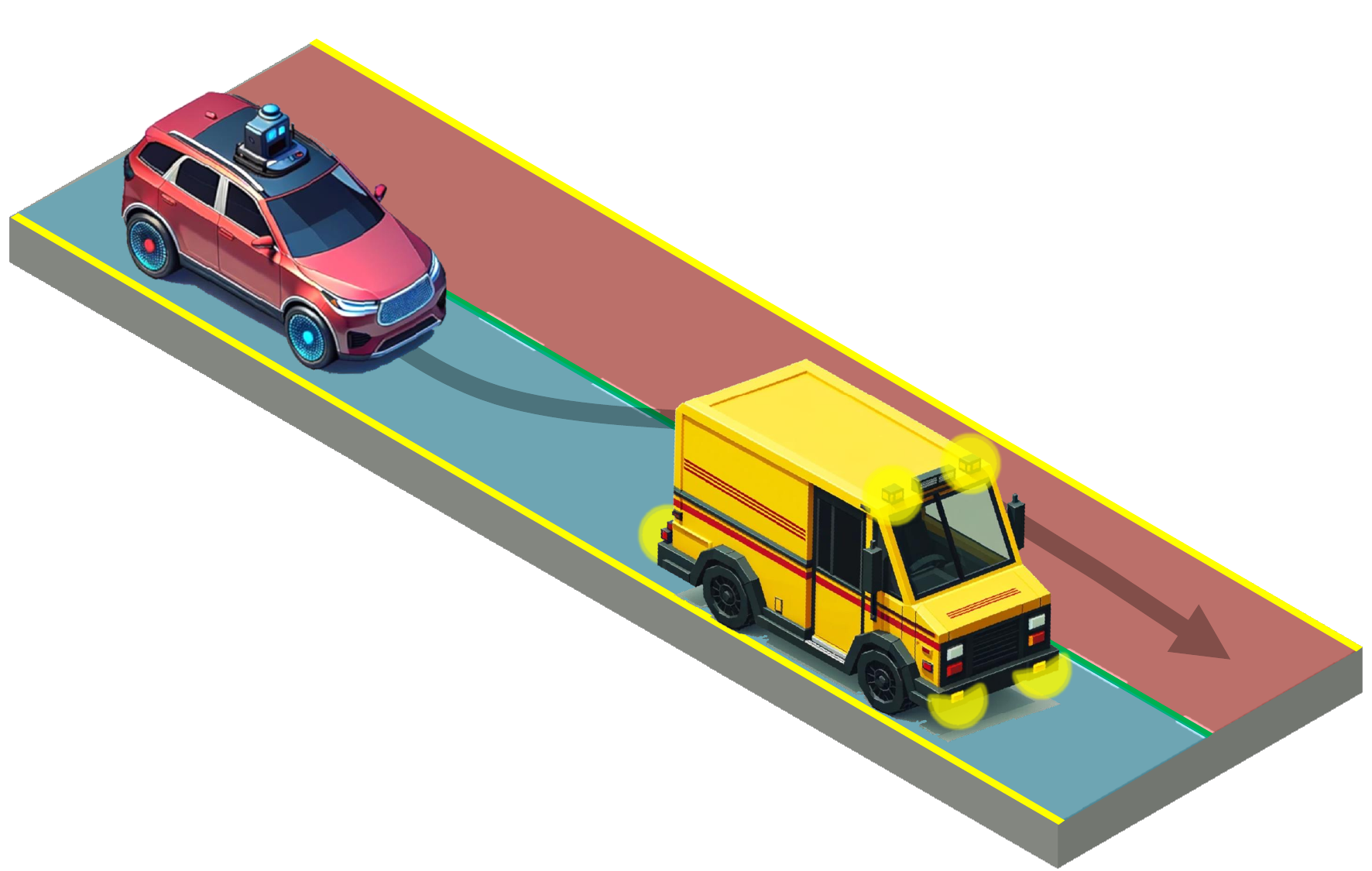}} &
        \subfloat[Scenario 2)]{\includegraphics[width=0.48\textwidth]{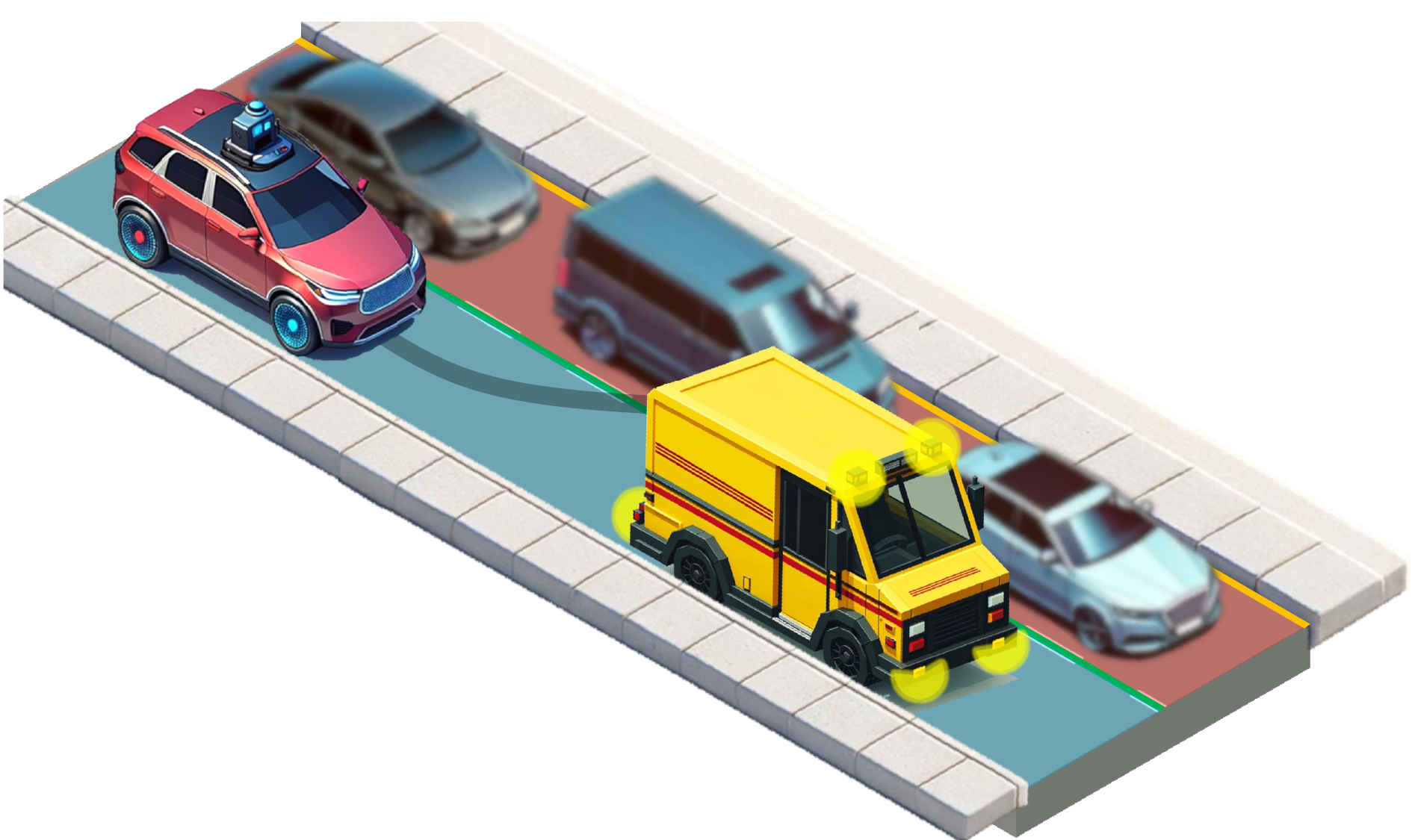}}
    \end{tabular}
    \caption{Visualization of the example scenarios. The red automated vehicle is approaching a yellow mail van blocking its lane. To continue its mission, it must perform a lane-change maneuver as visualized by the black example trajectory. Different infrastructure elements are highlighted as follows: The ego-lane of the automated vehicle is shown in blue, the adjacent lane with unknown driving direction is shown in red, broken white lane boundaries are marked in green, solid white lane boundaries are marked in yellow, and unmarked lane boundaries are marked in orange.}
    \label{fig:scenario}
    \vspace{-0.5em}
\end{figure*}

To summarize, the contributions of this paper are threefold: In Section~\ref{sec:capabilitygraph}, we introduce a scenario- and capability-based approach to the derivation of requirements for environment perception datasets. In Section~\ref{sec:scenario}, we demonstrate the efficacy of this approach in a small example scenario by deriving a set of necessary infrastructure elements for labeling lane detection datasets for mapless driving. To complete our discussion, in Section~\ref{sec:datasets}, we provide a comprehensive list of existing lane detection datasets and use the previously derived elements for a meaningful assessment and comparison, offering valuable insights into the current landscape of existing lane detection datasets and their application to mapless driving.

\section{Related Work}

Current practices in the development, documentation, and evaluation of lane detection datasets for automated driving lack a standardized approach. Existing datasets, as documented by their authors, typically emphasize general needs like scale and diversity, but often fall short in providing structured requirements, specific rationales for their approach to labeling, or comprehensive evaluations. Common evaluation metrics, while acknowledging the importance of dimensions such as weather, time of day, or scene type, tend to be overly simplistic and often reduce these complex variables to basic, sometimes binary categories~\cite{kitti:fritsch:2013,vpgnet:lee:2017,bdd100k:yu:2020,fiveaidataset:roberts:2018,apolloscape:huang:2019}.

Despite the critical role of dataset construction as the foundation of machine learning, this area remains under-researched in general~\cite{garbage:geiger:2020,genealogy:denton:2021,datasheets:2021:gebru,softwareengineering:hutchinson:2021}. Cha et al.~\cite{grindforgooddata:cha:2022} distill 19 interviews with machine learning experts into a six-step dataset construction pipeline, where they emphasize early-stage requirement identification in the first step. Vogelsang and Borg~\cite{requirementsml:vogelsang:2019} note the rising significance of data requirements in machine learning, suggesting the potential need for a distinct class of data requirements. Hutchinson et al.~\cite{softwareengineering:hutchinson:2021} propose a transparency framework for dataset development, drawing parallels with software engineering practices to stress the need for thorough documentation.

Literature specifically addressing lane detection datasets is limited. Shirke and Udayakumar~\cite{lanedatasetssurvey:shirke:2019} provide a comparison of six lane detection datasets within a review that includes eight datasets in total. However, their brief four-page survey uses only very general categories for comparison and lacks a structured, in-depth analysis. Other papers offer overviews of existing datasets within broader discussions on lane detection methods~\cite{ttlane:liang:2020,curvelane:xu:2020,bdd100k:yu:2020,lanemarkingsurvey:zhang:2021} or datasets for automated driving in general~\cite{safetodrive:guo:2019,addatasets:bogdoll:2022,osdataecosystem:li:2023,addatasets:liu:2024}. These comparisons typically focus on basic attributes like dataset size, resolution, and specific features like curve presence, but do not delve into the nuances necessary for a comprehensive dataset evaluation. For instance, they might note the inclusion of labels for lane detection without specifying if these labels include detailed type information for labeled lane boundaries~\cite{curvelane:xu:2020,bdd100k:yu:2020,safetodrive:guo:2019,addatasets:bogdoll:2022,osdataecosystem:li:2023,addatasets:liu:2024}. Moreover, even when the presence or absence of type information is noted, descriptions of the provided types are often omitted~\cite{ttlane:liang:2020,lanemarkingsurvey:zhang:2021}, which impedes the evaluation of particular datasets for specific applications.

In summary, while there is an emerging recognition of the need for more structured approaches in dataset development and evaluation, existing literature demonstrates gaps in both standardization and depth of analysis.

\section{Capability Graphs for Requirement Generation}
\label{sec:capabilitygraph}

\begin{figure}[t!]
    \vspace{0.2cm}
    \centering
    \includegraphics[width=\linewidth]{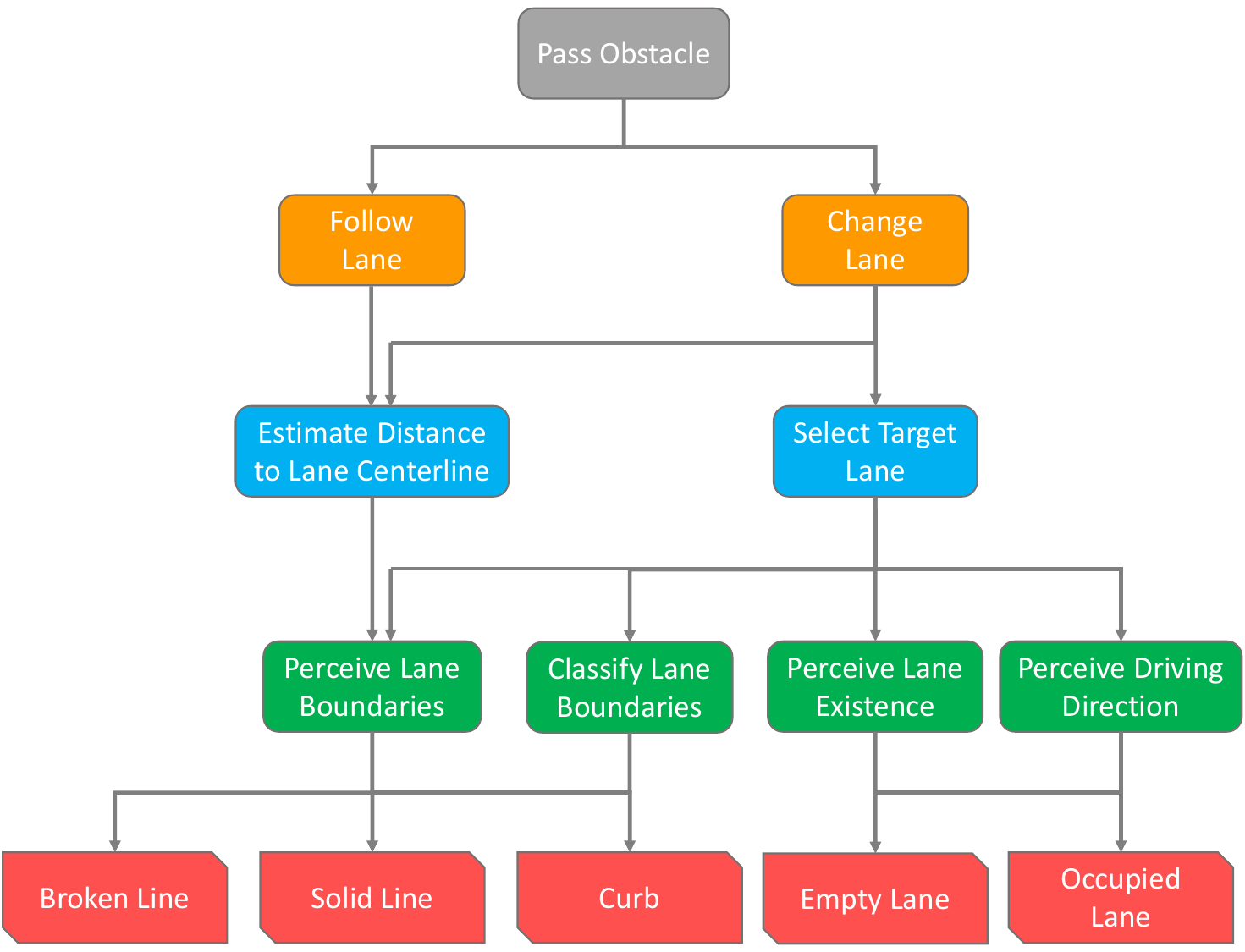}
    \caption{Visualization of a relevant subsection of the capability graph for the example scenarios with the overall mission goal in gray, visible external behavior in orange, high-level capabilities in blue, low-level capabilities in green, and relevant infrastructure elements in red.}
    \label{fig:capabilitygraph}
    \vspace{-0.5em}
\end{figure}

Given the identified need for more structured approaches, the creation of underlying datasets is recognized as a pivotal factor in the development of deep learning-based perception systems for automated driving. Developing effective perception systems and arguing the safety of the developed systems therefore necessitates a systematic derivation of dataset requirements, including a detailed specification of the characteristics of inputs and outputs. A significant step in this process is the identification of relevant objects for recording and labeling. This step is essential to ensure that perception modules based on the dataset can later effectively inform decision-making processes and support the driving task. Previous literature has proposed \emph{capability graphs} as a tool for the structured derivation of \emph{system requirements}, capturing the required properties to safely complete the dynamic driving task, as well as their dependencies~\cite{skillbaseddevelopment:nolte:2017}.

The use of capability graphs in the early stages of the development process for automated vehicles was originally proposed by Reschka et al.~\cite{skillgraph:reschka:2015}, based on earlier works of Maurer~\cite{emsvision:maurer:2000}, Pellkofer and Dickmanns~\cite{pellkofer:behavior:2002}, and Bergmiller~\cite{bergmiller:towardsfunctionalsafety:2015}. Reschka et al. use capability graphs to visualize the dependencies between different capabilities and system requirements during the concept phase as outlined in ISO~26262~\cite{iso26262:iso:2018}. This approach also aligns with the guidelines provided in Section 5, Specification and Design, of ISO~21448 (SOTIF)~\cite{iso21448:iso:2022}. To construct capability graphs, they begin with a maneuver-based description of the desired external behavior as defined in functional scenarios. By progressively detailing the necessary capabilities (which form the nodes of the graph) and their dependencies (which form the edges), a directed acyclic graph is created. This graph visually represents the connections between broad high-level capabilities and specific low-level capabilities, assisting in the aforementioned structured derivation of requirements~\cite{skillgraph:reschka:2015,bagschik:systemsperspective:2018}.

For learning-based systems, the resulting capabilities significantly depend on the datasets used during the training process. We therefore transfer the idea of using capability graphs for deriving system requirements to deriving requirements for datasets. Our methodology employs a layered approach that begins with a representative set of operational scenarios. It proceeds by partitioning these scenarios into individual maneuvers, each requiring specific capabilities. These are further refined within the construction of a capability graph, which serves as a methodical tool for mapping the transition from broad, high-level capabilities to specific, low-level ones. This structured approach facilitates the systematic derivation of detailed technical requirements and aids in the identification of distinct semantic elements that the perception system must perceive and distinguish. A more formal and comprehensive exploration of the capability graph development process is detailed in~\cite{automaticskillgraph:jatzkowski:2021}. This includes a discussion on how capability graphs could be automatically generated from representations of the Operational Design Domains (ODDs). An example of a capability graph is displayed in Figure~\ref{fig:capabilitygraph}.

We will next describe the example scenarios that we use in the remainder of this paper and which served as the foundation for this graph.

\section{Scenario Description and Requirement Derivation}
\label{sec:scenario}

Within the development process for automated vehicles, scenarios are introduced during the concept phase as a way to handle the complexities of real-world environments~\cite{scenarios:menzel:2018}. They can summarize large numbers of variations into humanly understandable descriptions that facilitate expert discussions, requirements engineering, and safety case development.

The scenarios we use as an example for our comparison of lane detection datasets are depicted in Figure~\ref{fig:scenario}. While a typical development process would feature the comprehensive exploration of a broad set of scenarios and the creation of a detailed scenario catalog, we focus on these two examples for clarity and brevity. They consider an automated vehicle approaching a parked mail van blocking its lane. As it approaches, the vehicle must determine the feasibility, legality, and safety of a lane change. To add additional detail to our analysis, we examine two variations of this basic scenario: The first contains clearly marked lanes and an empty adjacent lane, providing unobstructed visibility of the outer lane boundary. In the second scenario, the outer lane boundaries are delineated only by curb stones, while traffic in the adjacent lane occludes the road surface and outer lane boundary.

We assume that the main source of information regarding road characteristics is the vehicle's own environment perception system, and we will focus our analysis on this particular perception module. Other functions, like those for detecting and classifying road users, are outside the scope of this paper and are presumed to perform robustly within the example scenarios.

\begin{table*}[t]
    \vspace{0.2cm}
    \caption[Lane Detection Datasets - Overview]{Lane Detection Datasets - Overview\\Categories: ``Environment Urban'', ``Environment Highway'', ``Variation in Weather and Time of Day'', ``Contains Lane Boundaries'', ``Also Those Marked by Curb'', ``Also for Temporarily Occluded Areas'', ``Contains Class Information'', ``Contains Lane Areas'', ``Contains Road Area'', ``Contains Driving Direction''\\``Environment~Urban'':~$\circ$~only inner-city highways and large avenues, \cmark~also smaller roads\\``Lane~Boundaries'':~$\circ$~only ego-lane boundaries, \cmark~multiple/all lane boundaries\\``Driving~Direction'':~$\circ$~alternatively drivable areas share the same driving direction as the ego-lane}
    \label{tab:overview}
    \setlength\tabcolsep{3.45pt}
    \centering
    \begin{tabular}{l c r@{}l | c c | c | c c c c | c c c }
        \toprule
        Name \& Reference & Release Date & \multicolumn{2}{c}{\# Images} & \multicolumn{2}{c}{Environment} & Weather \& & Lane       & +Curb  & +Occluded & Classes & Lane  & Road & Driving\\
                          &              & &                             & Urb.    & Hwy.                  & Time Var.  & Boundaries &        &           &         & Areas & Area & Direc.\\
        \midrule
        Caltech Lanes~\cite{caltechlanes:aly:2008} & 06/2008 & 1,&225    & $\circ$ &                       &            & \cmark     &        &           & \cmark  &       &      & \\
        KITTI~\cite{kitti:fritsch:2013} & 10/2013 & &579                 & \cmark  &                       &            &            &        &           &         & ego only & \cmark & \\
        TuSimple~\cite{tusimple:2017} & 07/2017 & 6,&408                 &         & \cmark                &            & \cmark     &        &           &         &       &      & \\
        VPGNet~\cite{vpgnet:lee:2017} & 10/2017 & 20,&836                & \cmark  & \cmark                & \cmark     & \cmark     &        &           & \cmark  &       &      & \\
        ELAS~\cite{elas:berriel:2017} & 12/2017 & 16,&992                & $\circ$ & \cmark                &            & $\circ$    & \cmark &           & \cmark  &       &      & \\
        CULane~\cite{culane:pan:2018} & 12/2017 & 133,&235               & \cmark  & \cmark                & \cmark     & \cmark     & \cmark & \cmark    &         &       &      & \\
        BDD100k~\cite{bdd100k:yu:2020} & 05/2018 & 100,&000              & \cmark  & \cmark                & \cmark     & \cmark     & \cmark &           & \cmark  & ego+altern. & & $\circ$ \\
        Five AI~\cite{fiveaidataset:roberts:2018} & 07/2018 & 23,&979    & \cmark  & \cmark                & \cmark     &            &        & \cmark    &         & ego+parallel & \cmark & \\
        DET~\cite{det:cheng:2019} & 06/2019 & 5,&424                     & $\circ$ &                       &            & \cmark     &        &           &         &       &      & \\
        Unsup. LLAMAS~\cite{llamas:behrendt:2019} & 10/2019 & 100,&042   &         & \cmark                &            & \cmark     &        & \cmark    &         &       &      & \\
        Jiqing Expwy.~\cite{jiqingexpressway:feng:2019} & 12/2019 & 210,&610 &     & \cmark                &            & \cmark     &        &           &         &       &      & \\
        3D Lane Syn.~\cite{3dlanesynthetic:guo:2020} & 03/2020 & 7,&498  & $\circ$ & \cmark                &            & \cmark     &        & \cmark    & \cmark  & all   & \cmark & \\
        CurveLanes~\cite{curvelane:xu:2020} & 07/2020 & 150,&000         & \cmark  & \cmark                & \cmark     & \cmark     & \cmark & \cmark    &         &       &      & \\
        VIL-100~\cite{vil-100:zhang:2021} & 08/2021 & 10,&000            & $\circ$ & \cmark                &            & \cmark     &        &           & \cmark  &       &      & \\
        \bottomrule
    \end{tabular}
\end{table*}

We identify two key maneuvers within these scenarios, as shown in Figure~\ref{fig:capabilitygraph}: lane following and lane changing. Lane following requires that the vehicle estimates its distance to the lane center line with reasonable accuracy. In our simple scenario it is sufficient to recognize just one of the two lane boundaries to do this, though this is not always the case.

The lane-change maneuver is more complex, requiring the vehicle to assess the existence of an adjacent lane, allowed driving direction, dimensions, and the legality of the maneuver. Recognizing the outer boundary of an unoccupied lane is crucial, as it differentiates a vehicular lane from other infrastructure entities like bike lanes.

It is important to note that some characteristics, such as driving direction, could also be inferred in subsequent non-perception stages, for example, by analyzing the movement of other vehicles in the modeled vehicle environment. However, we argue that this inference should ideally be included as part of the perception tasks. Many environmental cues, like the overall visual appearance of a scene, are abstracted away in later processing steps but could be utilized by a deep learning-based perception system.

Based on these considerations, a perception system for our simple scenario would have to be able to reliably identify and classify features like solid and broken white lines, and curb stones, as shown in the bottom row of Figure~\ref{fig:capabilitygraph}. It should distinguish a regular traffic lane from similar looking infrastructure like bike lanes, bus lanes, or parking spots. The system must be able to infer lane existence and position when occluded by other traffic participants and predict the driving direction for both occupied and unoccupied lanes.

\section{Lane Detection Datasets}
\label{sec:datasets}

In this section, we will evaluate various lane detection datasets, based on the relevant road surface elements that we derived in the previous section. For this review, we define lane detection datasets as collections of near-ground level, non-aerial images of street scenes that provide either lane boundaries or lane areas. According to this definition, we include datasets such as TuSimple~\cite{tusimple:2017}, which offers polyline representations of lane boundaries, and KITTI~\cite{kitti:fritsch:2013}, which provides semantic segmentation masks of the ego lane and road area. However, we exclude datasets like Mapillary Vistas~\cite{mapillary:neuhold:2017}, which only contains pixel-wise segmentations of individual road markings, CeyMo~\cite{ceymo:jayasinghe:2022}, which contains road markings but does not feature lane boundaries, and object detection datasets such as Waymo~\cite{waymo:sun:2020}, even if they include maps of data acquisition areas, like nuScenes~\cite{nuscenes:caesar:2020}. This is due to the additional complexities associated with map projection onto sensor images which fall outside the scope of this paper, such as handling of foreground and background occlusions, and encoding of map elements.

For each dataset, we will provide general information and assess its utility in allowing a perception system to provide the necessary information for the two scenarios. A summary of the datasets under review is presented in Table~\ref{tab:overview}.

\subsection{Caltech Lanes}

Caltech Lanes~\cite{caltechlanes:aly:2008} was released in 2008 to enable automatic scoring of model-based lane detection techniques. It contains just 1,225 images captured on just two roads in Pasadena, California on a single clear and sunny day around noon.

It features annotations for all marked lane boundaries in the form of B\'{e}zier curves, classified into one of five classes, such as ``broken white'' or ``solid yellow''. However, it lacks labels for curbstones. A perception system trained on the Caltech Lanes dataset could therefore recognize all lane boundaries in the first scenario but would only detect a singular broken white line in the second scenario. It also lacks the concept of a lane and consequently annotations for driving directions. Together, these limitations would hinder the system's ability to fully understand lane dimensions or to facilitate safe lane changes.

\subsection{KITTI}

The widely recognized KITTI dataset, first introduced in 2012~\cite{kitti:geiger:2012} and subsequently expanded in 2013 to incorporate a road and lane detection benchmark~\cite{kitti:fritsch:2013}, is even smaller than the Caltech Lanes, containing only 579 images divided between training and test set. All images were captured under sunny conditions in and around Karlsruhe, Germany.

KITTI is distinguished by its approach of labeling the visible portion of the ``ego-lane'' and the ``road area'' through semantic segmentation masks, rather than annotating individual lane boundaries---an approach similar to the BDD100k dataset presented below. While this allows for effective lane keeping, its limitations include an inability to provide information on types of lane boundaries, the presence and size of adjacent lanes, or permissible driving directions, thereby restricting its usefulness for lane-changing assistance.

\subsection{TuSimple}

Released in 2017, the TuSimple dataset~\cite{tusimple:2017} features 6,408 images recorded on U.S. highways under sunny conditions, split into a training and test set. This makes it considerably larger than both the Caltech Lanes and KITTI datasets, yet it remains smaller than any of the other three datasets released in the same year. The dataset offers annotations for the location of up to five lane boundaries. However, it does not include class labels for lane boundaries or information on unmarked lanes. Foreground occlusions from other vehicles are disregarded, so that annotations are provided for lane boundaries even behind other vehicles, with some lines clearly constituting a best guess of the labeler.

A perception system trained on this dataset could support basic lane following functions even though it would only perceive the location of the central lane boundary in the second scenario. However, it lacks the necessary data for assessing the legality of lane changes since the dataset does not provide the type of lane boundaries. Furthermore, without the concept of a lane, it would be impossible for the system to identify if the space between two lane boundaries constitutes a physical lane or estimate its permissible driving direction.

\subsection{VPGNet}

Published alongside the Vanishing Point Guided Network in 2017, the VPGNet dataset~\cite{vpgnet:lee:2017} is notable for its size and variety, containing 20,836 images recorded over three weeks in Seoul, South Korea. It provides detailed annotations for visible road markings, categorized into seven types of lane boundaries and ten types for other road markings. However, there is no class for curbs or other unmarked lane boundaries.

Despite its comprehensiveness, it shares some limitations of the Caltech Lanes dataset, such as the neglect of unmarked lane boundaries like curbstones. While a perception system trained on the VPGNet dataset could identify the legality of crossing specific boundaries by the lane boundary type, it would fall short in providing information on the driving direction in adjacent lanes, which is crucial for executing safe lane changes.

\subsection{ELAS}

Introduced in 2017, the Ego-Lane Analysis System (ELAS) database~\cite{elas:berriel:2017} comprises roughly 17,000 images, captured exclusively during daytime and predominantly on multi-lane inner-city roads in and around the cities of Vit\'oria, Vila Velha, and Guarapari, Brazil. It only provides the position and class of the ego lane boundaries limited to a length of approximately 15 to 20 meters in front of the vehicle, which might restrict its utility at higher speeds. Notably, the ELAS dataset was, to our knowledge, the first lane detection dataset to provide a label and annotations for unmarked, faded or otherwise invisible lane boundaries, enhancing lane-keeping capabilities under these challenging conditions. However, the absence of information on areas beyond the immediate lane boundaries precludes more complex maneuvers, like lane changes, from being executed safely.

\subsection{CULane}

The CULane dataset~\cite{culane:pan:2018} was introduced in 2017 with the explicit goal of providing a large dataset for deep learning applications. Comprising a total of 133,235 images split into a training and test set, it remains one of the largest datasets. Images were captured by six vehicles in Beijing, China mostly on large multi-lane roads and inner-city highways under varying conditions. It includes annotations for up to four lane boundaries, even unmarked ones, allowing for the realization of a lane-keeping assist. However, it lacks classifications for lane boundaries and other critical information like driving directions, severely limiting its utility in supporting lane-change maneuvers.

\subsection{BDD100k}

The BDD100k dataset~\cite{bdd100k:yu:2020} released in 2018, is notable for its extensive collection of 100,000 video sequences, each 40 seconds in length. The image frame at the 10th second of each video is annotated for ten distinct tasks including lane detection and drivable area segmentation. It includes classifications for lane boundaries and markers in eight categories, notably including a ``road curb'' class. Additionally, it marks the current lane as directly drivable up to the first obstacle, with adjacent lanes marked as alternatively drivable, also up to the first obstacle. While all of this provides detailed information for lane following and lane-change maneuvers into empty lanes with the same driving direction, its limitations become evident in scenarios requiring information beyond the first obstacle or with opposite driving directions, which restricts its utility for safe lane-changing decisions.

\subsection{Five AI}

Launched in 2018, the Five AI dataset~\cite{fiveaidataset:roberts:2018} encompasses nearly 24,000 images from diverse locations across Great Britain. Similarly to KITTI, the Five AI dataset uniquely segments road areas into discrete lanes. Unlike KITTI, it includes all parallel and even occluded lanes. This approach has several advantages. It negates the need for converting lane boundaries into lane areas, eliminates ambiguities concerning the existence of lanes within lane boundaries, and enhances robustness against perception errors. However, the dataset omits specific labels for lane boundaries which poses challenges in determining the legality and safety of lane changes.

Although the dataset supports lane-keeping functions, even when lane boundaries are unclear or invisible, it falls short in assisting with lane-change maneuvers. The absence of lane boundary labels and indications of permissible driving directions means it cannot reliably determine the legality or safety of changing lanes.

\subsection{DET}

The DET dataset~\cite{det:cheng:2019}, released in 2019, is unique for using 5,424 black and white images captured with a Dynamic Vision Sensor (DVS). This sensor differs from traditional cameras by recording changes in light intensity at the pixel level, leading to ultra-fast response times. The authors compiled individual images by integrating data from the raw event stream over 30 milliseconds intervals. The dataset contains the location of the closest lane boundaries as pixel-wise semantic segmentation maps without specific class information. The treatment of occlusions and the handling of curbs are inconsistent, posing challenges for estimating the performance of a perception system in real-world scenarios.

While the dataset might be adequate for basic lane-keeping functions, it falls short in assisting with lane-change maneuvers, particularly due to insufficient information about adjacent lanes and the inconsistency in handling occlusions.

\subsection{Unsupervised LLAMAS}

The Unsupervised Labeled Lane Markers Using Maps (Unsupervised LLAMAS) dataset~\cite{llamas:behrendt:2019}, released in 2019, comprises 100,042 images sourced from 14 highway recordings across the United States. The dataset offers pixel-wise segmentation of individual road markings, but unlike the Mapillary Vistas or Apollo datasets, lane markings are grouped by lane boundaries. It shares limitations with other datasets like TUSimple and Jiqing Expressway, such as a narrow operational focus and absence of class labels for different types of road markings.

Like the TUSimple dataset, the provided information would allow for the design of a lane-keeping assist. A lane-change maneuver on the other hand would be inhibited by the lack of provided information in the first scenario and be entirely impossible in the second scenario.

\subsection{Jiqing Expressway}

By raw numbers the Jiqing Expressway dataset~\cite{jiqingexpressway:feng:2019}, published in 2019, stands as the largest currently available dataset. It consists of 40 video clips with 210,610 images recorded, as the name suggests, on the Jiqing Expressway in China on a single day in August around noon to early afternoon under largely clear, sunny skies and low traffic density, which greatly limits the diversity of the recorded scenes. All marked lane boundaries are labeled with simple poly-lines without class information.

Like TUSimple and Unsupervised LLAMAS, the provided information would allow for the design of a lane-keeping assist though the system would only provide the location of the central lane boundary in the second scenario. The lane-change maneuver on the other hand would be inhibited by the lack of provided information in the first scenario and be entirely impossible in the second scenario.

\subsection{3D Lane Synthetic Dataset}

The 3D Lane Synthetic Dataset~\cite{3dlanesynthetic:guo:2020}, published in 2020, is notable for its use of entirely computer-generated data, which comes at the advantage of providing precise labels even for occluded parts of the image. It also offers range images with the same size and format as the original color images. Disadvantages include a considerable gap between rendered and real data and substantial costs associated with crafting realistic synthetic images as the authors rely entirely on still images with hand-placed vehicles. The ground truth consists of lane boundaries and a set of lanes defined by lane boundaries. Lane boundaries and lanes are assigned a type from a list of four classes for lane boundaries like ``single solid'' or ``curb'' and three classes for lanes.

The availability of labels for marked lane boundaries and curb stones facilitates safe lane keeping. Lane-change maneuvers are enabled by lane boundary types and explicitly modeled lanes. An advantage of the simulated environment is the availability of accurate labels for occluded elements, which could allow the perception system to predict lane progression beyond the visible ground area based on contextual clues from the environment.

\subsection{CurveLanes}

The CurveLanes dataset~\cite{curvelane:xu:2020} was published in 2020 to correct the bias in existing datasets that predominantly feature straight roads. It comprises around 150,000 images, with over 90\% depicting curved roads in various Chinese cities, captured under diverse weather conditions and times of day. This makes it one of the largest and most varied datasets available.

Similar to other large datasets, size comes at the cost of labeled information. Like CULane, TuSimple, and the Jiqing Expressway dataset, CurveLanes only provides the location of lane boundaries, omitting details about boundary types. Nevertheless, locations are also provided for curb stones and occluded lane boundaries.

While this dataset supports reliable lane keeping, especially in scenarios with curved roads, it falls short in aiding lane-change maneuvers due to insufficient data on the type of lane boundaries and the characteristics of the road area beyond these boundaries.

\subsection{VIL-100}

VIL-100 (short for ``Video Instance Lane Detection-100'')~\cite{vil-100:zhang:2021} is a dataset for the detection of lane boundary instances in videos. It is one of the smallest datasets, comprising 100 videos with 100 frames, each collected mostly on highways and large inner-city avenues in China. It is noteworthy that the videos are created by splitting longer sequences. Different videos from the same sequence are then distributed across the test and training sets, failing to ensure the necessary independence of both sets.
In each frame all marked lane boundaries are labeled and classified into one of ten classes. However, it does not include labels for unmarked lane boundaries such as curb stones.

While the labeled information would enable effective lane keeping, the lack of information regarding driving direction or the existence of a lane would inhibit safe lane changes in the first scenario. Additionally, in the second scenario no information outside the current ego lane would be available precluding a safe lane change entirely.

\subsection{Summary}

The examination of various lane detection datasets provides critical insights into the current state of data availability and its suitability for automated driving applications, particularly in mapless scenarios.

Many datasets, such as TuSimple, KITTI, and CULane, offer valuable information for basic lane following functions. However, they often fall short in representing the diversity of real-world driving scenarios and there is a notable trade-off between dataset size and the quality and depth of labeling. Larger datasets, such as CULane and Jiqing Expressway, offer extensive data but often lack detailed information like lane boundary types and driving direction.
Smaller, more focused datasets like Caltech Lanes and VPGNet, though limited in size, provide more detailed annotations. Given that effective deep learning-based perception systems often require large datasets, there arises a need for improved tooling and more efficient processes to minimize manual efforts in dataset creation.

For complex driving maneuvers like lane changes, datasets need to provide more than just basic lane boundary information. This includes data on adjacent lanes, permissible driving directions, and types of lane boundaries.
Datasets such as BDD100k and Five AI, which provide segmented road areas and include adjacent lanes, partially meet these requirements. However, the absence of specific labels for lane boundaries and indications of permissible driving directions in the case of Five AI, as well as the handling of occlusions and lanes with opposite driving directions in the case of BDD100k limit their utility for safe lane-changing decisions. In contrast, the CULane and CurveLanes datasets attempt to offer accurate labels for occluded elements. However, these efforts are impeded by the lack of an objective ground truth behind obstructions. Projecting accurate map information into the sensor image could help to overcome this drawback.

In summary, there is a growing recognition of the need for datasets to encompass a wider range of real-world driving scenarios, but despite the wealth of information provided by some datasets reviewed here, no single dataset fully satisfies all the requirements for mapless automated driving. The development of datasets that accurately represent the diversity of real-world scenarios while at the same time providing comprehensive information for advanced automated driving tasks remains a key area for future research.

\section{Conclusion}
\label{sec:conclusion}

In this paper, we introduced a scenario- and capability-based approach as an initial step toward establishing structured and effective dataset development processes for mapless automated driving. Through the integration of capability graphs, this approach offers a systematic framework for identifying dataset requirements and ensuring that datasets align closely with the specific needs of automated driving technologies.

Our exploration into current lane detection datasets has revealed the pressing need for new datasets. Current datasets, while beneficial for certain aspects of lane detection, often do not provide the comprehensive and detailed information necessary for complex decision-making processes in mapless driving scenarios. This highlights an opportunity for future dataset development endeavors to focus on creating more nuanced and detailed datasets, encompassing a wide array of real-world scenarios and driving conditions.

While we concentrated on the scope and labeling aspects of these datasets, it is important to acknowledge that our study did not address other critical factors such as dataset bias, coverage, and labeling quality. These elements are crucial for assessing the overall effectiveness and reliability of datasets in real-world applications. The necessity for a comprehensive exploration of these aspects continues to represent a significant area for future research.

To conclude, the scenario- and capability-based approach presented in this paper marks a vital step toward more sophisticated dataset development processes in the realm of automated driving. However, it is just the beginning of a journey toward refining these processes and creating datasets that truly meet the evolving demands of this field. As the technology progresses, the continued advancement and innovation in dataset development processes will be critical to the success and safety of automated vehicles. The future of self-driving systems is contingent on our collective efforts to produce datasets that truly mirror the multifaceted nature of the real world.

\section*{Acknowledgment}

We would like to thank our colleague Robert Graubohm whose expertise and insightful feedback helped to refine the analysis and overall narrative of this paper. We would also like to thank our colleagues Marvin Loba and Nayel Salem whose valuable inputs enriched our research approach and findings.

%%%%%%%%%%%%%%%%%%%%%%%%%%%%%%%%%%%%%%%%%%%%%%%%%%%%%%%%%%%%%%%%%%%%%%%%%%%%%%%%

\bibliographystyle{bst/IEEEtran}
\bibliography{bib/IEEEabrv,%
              bib/dataset_bias,%
              bib/datasets,%
              bib/deep_learning,%
              bib/importance_of_data,%
              bib/scenarios,%
              bib/v_and_v_for_ml,%
              bib/autonomous_driving}

\end{document}